\documentclass{llncs}
\usepackage[table]{xcolor}
\usepackage{amsmath}
\usepackage{xspace}
\usepackage{mathtools}
\usepackage{algpseudocode}
\usepackage[ruled,vlined]{algorithm2e}

\usepackage{multirow}
\usepackage{graphicx}
\usepackage{color}
\usepackage{makeidx}  

\newcommand{\STATSEED}{Stat. Tree-based Population Seeding\xspace }
\newcommand{\STAT}{Stat. Tree\xspace}
\newcommand{\SHIFT}{Shift Buffer\xspace}
\newcommand{\RHEA}{Vanilla RHEA\xspace}

\begin{document}

\title{\ \\ \LARGE\bf  Statistical Tree-based Population Seeding for Rolling Horizon EAs in General Video Game  Playing}

 \author{Edgar Galv\'an\inst{1,2}, Oxana Gorshkova\inst{1},\\  Peter Mooney\inst{1,2}, Fred Valdez Ameneyro\inst{1,2} and Erik Cuevas\inst{2}}
\institute{Hamilton Institute, Maynooth University, Ireland
\and
Naturally Inspired Computation Research Group, \\Department of Computer Science, Maynooth University, Ireland\\
 \email{\{edgar.galvan,peter.mooney\}@mu.ie; \{oxana.gorshkova.2020, fred.valdezameneyro.2019\}@mumail.ie}
\and
 Centro Universitario de Ciencias Exactas e Ingenierías (CUCEI),\\ Universidad de Guadalajara, M\'exico\\
 \email{erik.cuevas@cucei.udg.mx}
}

%


\frontmatter          
\pagestyle{headings}  
\addtocmark{Hamiltonian Mechanics} 

\maketitle

\begin{abstract}
 Multiple Artificial Intelligence (AI) methods have been proposed over recent years to create controllers to play multiple video games of different nature and complexity without revealing the specific mechanics of each of these games to the AI methods. In recent years, Evolutionary Algorithms (EAs) employing rolling horizon mechanisms have achieved extraordinary results in these type of problems. However, some limitations are present in Rolling Horizon EAs making it a grand challenge of AI. These limitations include the wasteful mechanism of creating a population and evolving it over a fraction of a second to propose an action to be executed by the game agent. Another limitation  is to use a scalar value (fitness value) to direct evolutionary search instead of accounting for a mechanism that informs us how a particular agent behaves during the rolling horizon simulation. In this work, we address both of these issues. We introduce the use of a statistical tree that tackles the latter limitation. Furthermore, we tackle the former limitation by employing a mechanism that allows us to seed part of the population using Monte Carlo Tree Search, a method that has dominated multiple General Video Game AI competitions. We show how the proposed novel mechanism, called Statistical Tree-based Population Seeding, achieves better results compared to vanilla Rolling Horizon EAs in a set of 20 games, including  10 stochastic and 10 deterministic games.
\end{abstract}

\section{Introduction}
\label{sec:introduction}

General video game playing aims to produce Artificial Intelligence (AI) agents capable of playing a variety of games rather than playing a specific game. This is one of the grand challenges of AI. Games are excellent benchmark problems that  provide opportunities to test and analyse algorithms in AI. The General Video Game AI (GVGAI) framework~\cite{8664126} is a popular setting that offers a collection of games of different characteristics and this has been used extensively. 

Statistical forward planning techniques, such as Monte Carlo Tree Search (MCTS) \cite{DBLP:conf/ecml/KocsisS06} and Rolling Horizon Evolutionary Algorithms (RHEA) \cite{DBLP:conf/gecco/PerezSLR13}, are considered the dominant AI techniques in GVGAI competitions~\cite{8664126}. RHEAs have become real contenders to tree search-based methods that have dominated many GVGAI competitions for many years. \RHEA models sequences of actions encoded in  individuals that are improved through an evolutionary process. The algorithm can be stopped at any time and returns the first action of the fittest individual to be executed by the game agent. 

\RHEA ignores valuable information available when testing the individuals in the evolutionary process. This issue has been addressed with the use of a  statistical tree approach~\cite{10.1145/2739480.2754811}, the employment of a population seeding \cite{gaina2017rhseeding} and the use of a  shift buffer mechanism~\cite{8080420}, to mention a few. All these methods  yielding better results compared to \RHEA. However, most of the \RHEA enhancements waste computational power evaluating meaningless individuals and struggle to maintain a diverse population throughout the evolutionary process, which is important to avoid getting stuck in local optima. The lack of diversity in \RHEA is likely to happen given the small number of individuals forming a population due to the limited time budget available to the algorithm in which potential solutions can be evolved. 

The main contribution of this work is the proposal of novel mechanisms for \RHEA that exploits the theoretical guarantees of the Upper Confidence Tree (UCT) policy, explained in Section~\ref{sec:background} and heavily employed by MCTS. We use this to generate tree-based individuals and seed part of the evolved population saving valuable computational time. Valuable information gathered during the rolling horizon simulation of the EAs is used and is stored in an statistical tree.

The remainder of this paper is organised as follows. Section~\ref{sec:relatedWork} presents the related work. Section~\ref{sec:background} outlines the background in Monte Carlo Tree Search, Evolutionary Algorithms (EAs), Rolling Horizon EAs, and GVGAI. Section~\ref{sec:proposedApproach} discusses our proposed approach. Section~\ref{sec:experimental} presents the experimental setup used and Section~\ref{sec:results} presents and discusses the results obtained by all the approaches adopted in this study. Section~\ref{sec:conclusions} concludes this paper with some remarks.

\section{Related Work}
\label{sec:relatedWork}


Statistical tree enhancements, inspired by tree search algorithms, were devised by Perez et al.~\cite{10.1145/2739480.2754811}. Statistical tree enhancements targets some of the shortcomings of \RHEA of not using information knowledge obtained from previous game cycles.  The idea of this approach is to build a tree with the statistics of the scores acquired during the game simulations. The tree is built while evaluating each individual’s action plan. Each action is added as a node to the tree starting from the root. The root represents a node whose action was chosen on the previous game step. Each node contains information of how many times it has been visited, the action assigned to it and also an accumulated reward. This is called Open Loop Search and it means that the game states (generative models) are not stored in the nodes. This is  important because the game states can become obsolete and provide irrelevant information. The tree is built by evaluating the first action of the individual and this action is added as a child to the root node. All subsequent actions are added in the similar manner. When the last action is applied using a forward model, the game state is evaluated and its result is assigned to the individual's fitness. This value is then back-propagated to the root, incrementing the node visits count and augmenting stored rewards with the fitness value.

Some other interesting works have been proposed to improve  \RHEA. Gaina et al.~\cite{8080420} used four approaches to overcome some limitations present in \RHEA. The first approach, called   Bandit-based mutation, aims at using a bandit system for mutations at both individual and gene level. Interestingly, Gaina et al. used this to revert mutations when no improvement in the value of a given individual was observed. The second approach, named statistical tree, keeps statistical information in  game trees associated to each individual evolved. The fitness of each individual is employed to update the statistics stored in each node that has been visited during the evaluation of the individual. The third approach proposed by Gaina et al., called shift buffer,  is a simple yet effective technique that maximises the information gained during the limited time to make a decision. Instead of discarding information in every generation, the authors simply add a new random action at the end of each individual in the population.  The last approach called roll-outs, inspired by the work of Horn et al.~\cite{7860384} and Monte Carlo simulations, consists of rolling a simulation and then selecting a random selection of actions and game simulations using the forward model. The authors reported competitive results compared to tree-based methods. In particular, shift buffer, in combination with other approaches, such as roll-outs, yielded the most promising results.

More recently, Santos and Hauck~\cite{8636918} proposed two ideas combined into one approach: (i) one-step-look-ahead  and (ii) a  redundant action avoidance policy with the goal of overcoming some limitations present in  \RHEA. The former idea consists of using the forward model to try every possible action available from the current game state with the goal of selecting the best possible action. This was executed after shift buffer was applied, as proposed by Gaina et al.~\cite{8080420} and explained previously. The latter idea proposed by Santos and Hauck seeks to avoid redundant actions. More specifically an action is randomly selected whenever a set of sequence of actions is present in the individuals forming a population. They tested their approach using 20 different games and compared their results against \RHEA and \SHIFT obtaining better results overall.

\section{Background}
\label{sec:background}

\subsection{Monte Carlo Tree Search}
\label{sec:mcts}

Monte Carlo Tree Search (MCTS) is a sampling method for finding \textit{optimal decisions} by performing random samples within the decision space and building a tree according to the partial results.  These Monte Carlo methods work by approximating future rewards that can be achieved through random samplings. The evaluation function of MCTS relies directly on the outcomes of simulations. The accuracy of this function increases by adding more simulations, thus the optimal search tree is guaranteed to be found with infinite memory and computation~\cite{DBLP:conf/ecml/KocsisS06}. However, in more realistic scenarios (e.g., limited computing resources available), MCTS can produce very good approximate solutions. MCTS has gained a lot of popularity thanks to its recent success in the board game of computer Go~\cite{gelly2006}, where the space of solutions is $10^{170}$ and up to 361 legal moves. A problem, that until recently, was highly difficult for  AI, and considered much more harder than chess, and where MCTS has obtained excellent results, including achieving dan -- master -- level at the 9 x 9 board game. This is, perhaps, the reason why MCTS has been used heavily in two-player board games. 

However, researchers have begun to apply MCTS to other research problems. For instance, MCTS has been explored in  constraint satisfaction (e.g.,  constraint problems~\cite{DBLP:conf/ijcai/BabaJIY11}) and energy-based problems~\cite{6850585,DBLP:conf/iisa/LopezLPCC14}.

\subsubsection{The Mechanics Behind MCTS}

MCTS relies on two key elements: (a) that the true value of an action (in our problems, an action could be turn right, turn left, and so on) can be approximated using simulations; and (b) that these values can be used to adjust the policy towards a best-first strategy.  The algorithm, explained later in this section, builds a partial tree, guided by the results of previous explorations of that tree.  The algorithm iteratively builds a tree until a condition is reached or satisfied (e.g., number of simulations, time given to perform Monte Carlo simulations), then the search is halted and the best performing action is executed. In the tree, each node represents a state, and directed links to child nodes represent actions leading to subsequent states. 

Like many AI techniques, MCTS has several variants. Perhaps, the most accepted steps involved in MCTS are those described in~\cite{DBLP:conf/aiide/2008} and are outlined as follows: (a) \textit{Selection:} a selection policy is recursively applied to descend through the built tree until an expandable node has been reached. A node is classified as expandable if it represents a non-terminal state, and also, if it has unvisited child nodes;  (b) \textit{Expansion:} normally one child is added to expand the tree subject to available actions; (c) \textit{Simulation:} from the new added nodes, a simulation is run to obtain an outcome (e.g., reward value); and (d) \textit{Back-propagation:} the outcome from the simulation step is back-propagated through the selected nodes to update their statistics.


Simulations in MCTS start from the root state (in our case, from the current time when an action for an electric device should be made) and are divided in two stages. When the state is added in the tree, a \textit{tree policy} is used to select the actions (the selection step is a key element and it is discussed in detail later in this section). Otherwise, a \textit{default policy} is used to roll out simulations to completion.  One element that contributed to enhance the efficiency in MCTS is the selection mechanism proposed in~\cite{DBLP:conf/ecml/KocsisS06}. The main idea of this proposed selection mechanism was to design a Monte Carlo search algorithm that had a small probability error if it were stopped prematurely and eventually converged to the optimal solution given enough time. The selection mechanism nicely balances exploration \textit{vs.} exploitation and this will be explained in the following paragraphs.

\subsubsection{Upper Confidence Bounds for Trees}

As indicated previously, MCTS works by approximating ``real'' values of the actions  that may be taken from the current state. This is achieved through building a search or decision tree. The success of MCTS depends heavily on how the tree is built and the selection process plays a fundamental role in this.  One particular selection mechanism that has proven to be very reliable is the UCB1~\cite{10.1023/A:1013689704352}. Formally, UCB1 is defined as:

\begin{equation}
UCB1 = \bar{X_j} + 2K \sqrt{\frac{2 \cdot \ln(n)}{n_j}} 
  \label{eq_UCT}
\end{equation}

\noindent where $n$ is the number of times the parent node has been visited, $n_j$ is the number of times child $j$ has been visited, $\bar{X_j}$ is the mean reward of the node $j$ and $K > 0$ is a constant. In case of a tie for selecting a child node, a random selection is normally used~\cite{DBLP:conf/ecml/KocsisS06}. Thus, this selection mechanism is successful due to its emphasis on balancing both exploitation (first part of Eq.~\ref{eq_UCT}) and exploration (second part of Eq.~\ref{eq_UCT}). Every time a node is visited, the denominator of the exploration part increases resulting in decreasing its overall contribution. If, on the other hand, another child node of the same parent node is visited, the numerator increases, so the exploration values of unvisited children increase.  The exploration term in Eq.~\ref{eq_UCT} guarantees that each child node has a selection probability greater than zero, which is essential given the random nature of the play-outs.

\subsection{Evolutionary Algorithms}
\label{sec:eas}
Evolutionary Algorithms (EAs)~\cite{EibenBook2003} refer to a set of stochastic optimisation bio-inspired algorithms that use evolutionary principles to build robust adaptive systems. EAs work with a population of $\mu$-\textit{encoded} potential solutions to a particular problem. Each potential solution, commonly known as an individual, represents a point in the search space, within which the optimum solution lies. The population is evolved by means of genetic operators, over a number of generations, to iteratively produce better results to the problem.  Each member of the population is evaluated using a fitness function to determine how good or bad the potential solution is for the problem at hand. The fitness value assigned to each individual in the population probabilistically determines how successful the individual will be at propagating (part of) its code to further generations. Better performing solutions will be assigned higher values (for maximisation problems) or lower values (for minimisation problems). The evolutionary  process  is repeated until a stopping condition is met. Normally this is until a maximum number of generations has been executed. The population in the last generation  is the result of exploring and exploiting the search space over a number of generations. It contains the best evolved potential solutions to the problem and may also, in some  cases, represent the globally optimum solution.  EAs have been successfully used in a wide variety of problems including games~\cite{5586508,DBLP:conf/cig/LopezO09,DBLP:conf/evoW/LopezSOB10}, energy-based problems~\cite{galvan_neurocomputing_2015,10.1007/978-3-319-48506-5_9,Galvan_SAC_2014,7529311}, evolvable hardware~\cite{galvan-Lopez2008,DBLP:conf/eurogp/LopezPC04}, file type detection~\cite{DBLP:conf/eurogp/KattanLPO10}, classification of imbalanced datasets~\cite{10.1145/3321707.3321854,Galvan-Lopez2016,10.1007/978-3-319-62428-0_22}, to mention a few examples.

\subsection{Rolling Horizon Evolutionary Algorithms}


Rolling Horizon Evolutionary Algorithms (RHEAs)~\cite{10.1145/2463372.2463413} are inspired by MCTS as explained in Section~\ref{sec:mcts}. RHEAs employ roll-outs and a generative model (a simulator) allowing the algorithm to see into the future for a short period of time (generally for a fraction of a second).  Using this information, the agent (individual) evolves a plan by means of EAs, explained in Section~\ref{sec:eas}, executes an action on the problem by firing the first action of its plan. This continues until a stopping condition is met. In the context of GVGAI, RHEAs are executed until the game is over. Details on how RHEA works can be found in greater detail within~\cite{10.1145/2463372.2463413}. 

\subsection{General Video Game Artificial Intelligence}
\label{sec:gvgai}
In this work, we use the General Video Game AI (GVGAI) framework~\cite{8664126} to test our proposed approach, called Statistical Tree-based Population Seeding (see Section~\ref{sec:proposedApproach}). The GVGAI framework provides a corpus of games, as well as means for creating games, that can be used as testbeds. The corpus consists of two-player games  and single-player 2D games of different genres, including  shooter, maze, survival, puzzle, and so on.  All games have their own rules, scoring and winning criteria. For example, in the game ``Aliens'', the agent (or controller)  wins when it kills all of the aliens and it gets points for every killed alien and an obstacle destroyed.  A controller (agent) in the game ``Survive Zombies''  wins if it is able to stay alive by avoiding being eaten by zombies for a particular amount of time. The avatar in this game has a health bar that decreases every time a zombie catches a player. Collecting honey helps restore the health bar. Furthermore, there are obstacles in the player’s way that must be avoided whilst also running away from zombies.

Different games have different actions that are legal for a controller to execute. In total, there are six actions in the GVGAI framework: left, right, up, down, nil, use and escape. In the planning track of the framework, controllers have access to the current game state; they can copy it and execute actions on the copy by using a forward model, necessary to perform roll-outs, thus simulating the game. Games can be deterministic or stochastic in nature. In stochastic games, a forward model can produce different future game outcomes from the current game state with each simulation, contrary to deterministic games.  All games can be played on five levels. Each level can be played independently, there is no connection between levels or on how the controller performs, as there is no requirement for a controller to win a  level to advance to the next level. A higher level can increase a complexity of the game played and can introduce new challenges to the player, and so on.  In our experiments, a controller has a budget of 900 forward model calls (equivalent to  40ms) to simulate the game and decide on an action to fire. This is aligned to the GVGAI competitions where controllers face disqualification if they exceed the time allowed for decision making.

\section{Statistical Tree-based Population Seeding}
\label{sec:proposedApproach}
We propose a novel mechanism that partly seeds the population of the RHEA using the UCB1 policy on a statistical tree. The statistical tree introduced in \cite{10.1145/2739480.2754811} is a RHEA improvement that constructs a tree with each state visited when individuals in the population are evaluated. Each state is modelled as a node and is linked to the previous state by an action. Each node stores two values: the times it has been visited and the cumulative reward. After each individual is evaluated the rewards are back-propagated from the corresponding leaf node to the root. In this way, the statistical tree keeps track of the rewards found by the RHEA's evolutionary process. RHEA returns the first action of the fittest individual as the action to be performed by the controller. In the statistical tree, the node linked to the root with the performed action is transformed into the new root node. Any nodes not part of the sub-tree beginning from the new root node are pruned. 

When the RHEA algorithm is required again to return a new action, the initial population is seeded with the statistical tree. The first individual is generated following the path in the tree leading to the best reward. The rest of the population is generated following the UCB1 policy from the root node. This action is repeated $m-1$ times, where $m$ is the population size, to generate the remaining $m-1$ individuals. It is important to evaluate each new individual before generating a new one to update the statistical tree.  The available computational power to make a decision is limited (900 forward model  calls) and as a result of this, the size of the population must be small. This can hinder the population's diversity with the consequence of individuals getting stuck in local optima. To alleviate this, the proposed algorithm attempts to inject a new individual after each generation. The individual is generated following the UCB1 policy on the statistical tree, and replaces the worst individual in the current population only if the newly tree-based generated individual performs better. Regardless of it being added to the population or discarded, the newly generated individual contributes to the construction and statistics of the statistical tree. This step helps to keep interesting individuals emerging in the within population while guiding the search towards promising regions.



\section{Experimental Setting}
\label{sec:experimental}
\subsection{Controllers}

We compare our proposed approached, dubbed (a) Statistical Tree-based Population Seeding, against (b) Vanilla RHEA~\cite{10.1145/2463372.2463413}. Furthermore, we also use (c) the Statistical Tree approach~\cite{10.1145/2739480.2754811}, and (d) the Shift Buffer mechanism~\cite{8080420}. The latter three approaches (b -- d), explained previously, heavily inspired our proposed approach and show how by borrowing elements from these and integrating new elements such as partially seeding a population, it is possible to get, overall, better results compared to these three approaches.

\subsection{Games}

\begin{table} [th]
\caption{List  of the 20 games used in our studies, including their names with their corresponding indexes divided in two categories: stochastic (G1 -- G10)  and deterministic (G11 -- G20).}

\centering
\resizebox{0.5\columnwidth}{!}{
\begin{tabular}{|c|l|c|l|}\hline
Index & Stochastic & Index & Deterministic \\ \hline
G1  & Aliens    & G11  &  Bait \\ \hline
G2 & Butterflies & G12 &  Camel Race \\ \hline
G3 & Chopper     & G13 &  Chase \\ \hline
G4 & Crossfire   & G14 &  Escape \\ \hline
G5 & Dig Dug    & G15 & Hungry Birds \\ \hline
G6 & Infection & G16 & Lemmings \\ \hline
G7 & Intersection & G17 & Missile Command \\ \hline
G8 & Roguelike    & G18 & Modality \\ \hline
G9 & Sea Quest    & G19 &  Plaque Attack \\ \hline
G10 & Survive Zombies & G20 & Wait for Breakfast \\ \hline
\end{tabular}
}
\label{tab:setOfGames}
\end{table}

Each of the  four  algorithms (controllers) used in this study were run on 20 games of the GVGAI corpus, on all 5 levels, 40 times each. Each algorithm had to play 4000 times, resulting in 16000 independent runs in total. To perform this large number of simulations, we used a supercomputer with  336 nodes where each node has 2x 20-core 2.4 GHz Intel Xeon Gold 6148 (Skylake) processors, 192 GiB of RAM. Furthermore, to provide a sound comparison of results, we selected ten stochastic games and ten deterministic games. The selection of these is based on performing  a uniform sampling from the list of games in Table~\ref{tab:setOfGames}.  The selection of these games is the result of analysing dozens of games based on the performance of MCTS expressed in terms of the winning rate for each of the 62 games used in Nelson~\cite{7860443} and based on the similarity of game features among 49 games studied in Bontager et al.~\cite{Bontrager2016MatchingGA}. 

\subsection{EAs Configuration and Parameters}

The experiments were conducted using a generational approach. The population size was set to 10 and the length of the individual (action plan) set to 14. All four algorithms used in this study used uniform crossover and mutation operators. Tournament selection (size 3) was used as a selection mechanism in our algorithms. Furthermore, we used elitism (1 individual) to ensure that the best individual was carried over to the next generation. Our algorithms were not limited in the number of generations. Instead, we set a budget of 900 forward model calls  every time an action was fired by any of the algorithms used in this study. We use the score as the fitness function in each of the four controllers used in this study.


\section{Results and Analysis}
\label{sec:results}

All results were converted into Formula-1 (F1) ranking scheme. That is, for each game, controllers are compared by their mean win rate. The controller with the highest average win rate gets 25 points, the second best controller is granted 18 points, and the remaining two controllers  get 15 and 12 points respectively. If two or more controllers achieved the same number of victories then they are compared by their scores. The overall results are displayed in Table~\ref{tab:f1_results}. A breakdown of the results for each game is presented in Table~\ref{tab:resultsStochastic} for stochastic games and  Table~\ref{tab:resultsDeterministic} for deterministic games. \RHEA is ranked 4th and is the worst of all the controllers used in this work. Our proposed Statistic Tree-based Seeding Population  algorithm ranked first in 10 games as well as first overall. It is closely followed by the \STAT variant. Both methods were tied in F1 scores in deterministic games, but \STATSEED had a better performance in stochastic games, showing that the proposed method has a potentially greater impact in the latter types of games.

\begin{table}[th]
\caption{F1 total points achieved by each controller used in this study.}
\centering
\resizebox{0.65\columnwidth}{!}{
  \begin{tabular}{|l|l|r|r|r|r|}\hline

Rank & RHEA Approach& F1 Points &  Avg. Points \\ \hline
 
1st &  \STATSEED &  412 & 49.67\% \\ \hline
2nd & \STAT & 382 & 48.37\% \\ \hline
3rd & \SHIFT &  315 & 45.92\% \\ \hline
4th & \RHEA & 291 & 46.77\% \\ \hline
  \end{tabular}
  }
\label{tab:f1_results}
\end{table}

\begin{table}[th]
\caption{F1 ranking of the controllers in the 10 \underline{stochastic} games. The darker the cell, the better the result.}
\centering
\resizebox{0.95\columnwidth}{!}{
\begin{tabular}{|l|*{10}{m{0.6cm}|}c|}\hline
RHEA Methods & G1 &      G2&      G3&      G4&      G5&      G6&      G7&      G8&      G9&      G10& Total \\ \hline
\STATSEED  & \cellcolor[gray]{.75}   15&      \cellcolor[gray]{.75}15&      \cellcolor[gray]{0}  \textcolor{white} {25} &      \cellcolor[gray]{0}  \textcolor{white} {25}&      \cellcolor[gray]{.75}15&\cellcolor[gray]{.75}      15&      \cellcolor[gray]{0}  \textcolor{white} {25}&      \cellcolor[gray]{.5}18 &      \cellcolor[gray]{0}  \textcolor{white} {25}   &   \cellcolor[gray]{0}  \textcolor{white} {25} & \cellcolor[gray]{0}  \textcolor{white} {203} \\ \hline
\STAT &\cellcolor[gray]{.5}        18&     \cellcolor[gray]{0}  \textcolor{white} {25} &     \cellcolor[gray]{0.9}12&      \cellcolor[gray]{.75}15 &     \cellcolor[gray]{.5}18 &     \cellcolor[gray]{0}  \textcolor{white} {25}  &    \cellcolor[gray]{.75}15&\cellcolor[gray]{.75}      15 & \cellcolor[gray]{.75}    15  &    \cellcolor[gray]{.75}15  & \cellcolor[gray]{.75}173\\ \hline
\SHIFT     & \cellcolor[gray]{0}  \textcolor{white} {25}  &    \cellcolor[gray]{.5}18 & \cellcolor[gray]{.5}    18  &    \cellcolor[gray]{0.9}12 &     \cellcolor[gray]{0}  \textcolor{white} {25}   &   \cellcolor[gray]{.5}18  &    \cellcolor[gray]{0.9}12  &    \cellcolor[gray]{0}  \textcolor{white} {25}  &   \cellcolor[gray]{0.9}12  & \cellcolor[gray]{0.9}   12  & \cellcolor[gray]{.5}177\\ \hline
\RHEA     & \cellcolor[gray]{0.9}      \cellcolor[gray]{0.9}12  & \cellcolor[gray]{0.9}   12&     \cellcolor[gray]{.75} 15   &   \cellcolor[gray]{.5}18&      \cellcolor[gray]{0.9}12   & \cellcolor[gray]{0.9}  12 &     \cellcolor[gray]{.5}18  &    \cellcolor[gray]{0.9}12   &   \cellcolor[gray]{.5}18 & \cellcolor[gray]{.5}     18 & \cellcolor[gray]{0.9}147 \\ \hline
\end{tabular}
}
\label{tab:resultsStochastic}
\end{table}

\begin{table}
\caption{F1 ranking of the controllers in the 10 \underline{deterministic} games. The darker the cell, the better the result.}
\centering
\resizebox{0.95\columnwidth}{!}{
\begin{tabular}{|l|*{10}{m{0.6cm}|}c|}\hline
RHEA Methods & G11 &      G12&      G13&      G14&      G15&      G16&      G17&      G18&      G19&      G20 & Total\\ \hline
\STATSEED    &	\cellcolor[gray]{0}\textcolor{white} {25}&	\cellcolor[gray]{0}\textcolor{white} {25}&	\cellcolor[gray]{.5}18&	\cellcolor[gray]{0}\textcolor{white} {25}	&\cellcolor[gray]{.5}18&	\cellcolor[gray]{0.9}12&	\cellcolor[gray]{.5}18&	\cellcolor[gray]{.5}18&	\cellcolor[gray]{0}\textcolor{white} {25}&	\cellcolor[gray]{0}\textcolor{white} {25} & \cellcolor[gray]{0}\textcolor{white}{209}\\ \hline
\STAT &	\cellcolor[gray]{.5}18&	\cellcolor[gray]{.5}18& \cellcolor[gray]{0}\textcolor{white} {25}&	\cellcolor[gray]{0.9}12	& \cellcolor[gray]{0}\textcolor{white} {25}&	\cellcolor[gray]{0}\textcolor{white} {25}&	\cellcolor[gray]{0}\textcolor{white} {25}&	\cellcolor[gray]{0}\textcolor{white} {25}&	\cellcolor[gray]{.5}18&	\cellcolor[gray]{.5}18 & \cellcolor[gray]{0}\textcolor{white}{209}\\ \hline
\SHIFT	& \cellcolor[gray]{0.9}12 &	\cellcolor[gray]{.75}15&	\cellcolor[gray]{0.9}12&	\cellcolor[gray]{.75}15&	\cellcolor[gray]{0.9}12&	\cellcolor[gray]{.5}18&	\cellcolor[gray]{.75}15&	\cellcolor[gray]{0.9}12&	\cellcolor[gray]{0.9}12&	\cellcolor[gray]{.75}15 & \cellcolor[gray]{.9}138\\ \hline
\RHEA	&  \cellcolor[gray]{.75}15&	\cellcolor[gray]{0.9}12&	\cellcolor[gray]{.75}15&	\cellcolor[gray]{.5}18&	\cellcolor[gray]{.75}15&\cellcolor[gray]{.75}	15&	\cellcolor[gray]{0.9}12&	\cellcolor[gray]{.75}15&	\cellcolor[gray]{.75}15&	\cellcolor[gray]{0.9}12 & \cellcolor[gray]{.75}144\\ \hline

\end{tabular}
}
\label{tab:resultsDeterministic}
\end{table}

The Mann-Whitney U test at 5\% significance level was used to compare controllers with each other to determine if the difference in scores and wins is statistically significant. Table ~\ref{tab:statisticalWin} shows the count of stochastic and deterministic games where the algorithm in the row significantly outperforms the algorithm in the column in win rates.  Table~\ref{tab:statisticalScores} is similar to this but now considers those significant scores between any two algorithms. Table~\ref{tab:detailsWins}, listed in the Appendix, shows the details of wins for each game attained by each of the four methods used in this work. The average scores are not shown due to space constraints. Based on these tables, it can be seen that both statistical tree-based  RHEA versions achieved the highest number of  games where they outperformed the other two algorithms. Our proposed approach, Statistical Tree-based Population Seeding,  had a significantly superior score in 13 games in total, and significantly superior win rates in 9 games, compared to the other three algorithms (controllers). The three games in which the Statistical Tree-based Population Seeding  significantly outperformed the \RHEA in win rates are \textit{Wait for breakfast}, \textit{Missile command} and \textit{Chopper}. Regarding scores, our approach  was significantly better than \RHEA in \textit{Wait for breakfast}, \textit{Roguelike}, \textit{Crossfire}, \textit{Chopper} and \textit{Intersection}. \textit{Missile command} is the only game in which both the Statistical Tree-based Population Seeding and the \STAT controllers significantly improve over the \RHEA in both scores and win rates. \textit{Dig-dug} is the only game in which the \STATSEED performed significantly worse than the \RHEA regarding scores. This is not the same situation for win rates. \STATSEED also struggled with the game $Lemmings$. One possible reason is that exploration in this game can be extremely costly and rewards are sparse. The exploratory nature of \STATSEED ends up actually impacting it negatively. In general, \STATSEED is a promising approach that improves over the \RHEA, being clearly superior to the \SHIFT and more robust than the \STAT in many cases.


\begin{table} [ht]
\caption{Count of games where the controller in the row has significantly \underline{better win rates} than the controller in the column.}
\centering
\resizebox{0.85\columnwidth}{!}{
\begin{tabular}{|l|c|c|c|c|}\hline
& Stat. Tree-based Seeding & \STAT & \SHIFT & \RHEA  \\  \hline
Stat. Tree-based Seeding &  -- & 2 & 4 & 3 \\ \hline
\STAT & 2 & -- & 3 & 3 \\ \hline
\SHIFT & 0 & 0 & -- & 0 \\ \hline
\RHEA &0 &  2 & 1 & -- \\ \hline
\end{tabular}
}
\label{tab:statisticalWin}
\end{table}

\begin{table} [ht]
\centering
\caption{Count of games where the controller in the row has significantly \underline{better scores} than the controller in the column.}
\resizebox{0.85\columnwidth}{!}{
\begin{tabular}{|l|c|c|c|c|}\hline
& Stat. Tree-based Seeding & \STAT & \SHIFT & \RHEA  \\  \hline
Stat. Tree-based Seeding &  -- & 3 & 4 & 6 \\ \hline
\STAT & 4 & -- & 4 & 5 \\ \hline
\SHIFT & 2 & 0 & -- & 5 \\ \hline
\RHEA & 1 &  1 & 0 & -- \\ \hline
\end{tabular}
}
\label{tab:statisticalScores}
\end{table}

\section{Conclusions}
\label{sec:conclusions}
In this paper we proposed a novel mechanism named Statistical Tree-based Population Seeding to be employed in Rolling Horizon Evolutionary Algorithms (RHEAs) with the goal of overcoming some limitations of the vanilla RHEA. This mechanism uses a Statistical Tree that keeps track of information gathered by the individuals contained in the population. Each of these individuals contribute to the information stored in this Statistical Tree every time a potential solution is rolled in a game simulation. We then use this tree to generate an individual to be seeded in the population in order to have diversity in the population as well as speeding up the process of evolution required in the limited time (a fraction of a second) to `fire' an action to be executed by the game agent. We demonstrated how our approach is significantly superior to \RHEA as well as being better, in average, to the other two controllers, named Statistical Tree and Shift Buffer, used in this study.  

\section*{Acknowledgements}
This publication has emanated from research conducted with the financial support of Science Foundation Ireland under Grant number 18/CRT/6049. The authors wish to acknowledge the DJEI/DES/SFI/HEA Irish Centre for High-End Computing (ICHEC) for the provision of computational facilities and support.

\bibliographystyle{abbrv}
\bibliography{rollingStats_final}

\section*{Appendix}

\begin{table} [ht]
\caption{Mean win rate achieved by each controller in each of the 20 games used in this study.}
\centering
\resizebox{0.95\columnwidth}{!}{
\begin{tabular}{|l|c|c|c|c|}\hline

  \multirow{2}{*}{Game name}	& \multicolumn{4}{c|}{Mean wins ($\pm$ standard deviation)}                 \\  \cline{2-5}
  & Stat. Tree-based Seeding & \SHIFT & \STAT  & \RHEA \\ \hline
aliens &	1.00 ($\pm$ 0.00) 	& 1.00 ($\pm$ 0.00) & 1.00 ($\pm$ 0.00) & 1.00 ($\pm$ 0.00)  \\ \hline
  bait	& 0.11 ($\pm$0.31) & 0.04 ($\pm$0.20)  & 0.09 ($\pm$0.28) & 0.07 ($\pm$0.25)\\ \hline
  butterflies & 0.92 ($\pm$0.27) & 0.92 ($\pm$0.26) & 0.98 ($\pm$0.14) & 0.88 ($\pm$0.32)\\  \hline
  camelRace	& 0.05 ($\pm$0.21) & 0.03 ($\pm$0.18) & 0.04 ($\pm$0.20)& 0.03 ($\pm$0.17)\\ \hline
  chase &	0.03 ($\pm$0.18) & 0.03 (0.17) & 0.05 ($\pm$0.21) & 0.035 ($\pm$0.18)\\ \hline    
chopper &	1.00 ($\pm$ 0.00) & 0.99 (0.09)	& 0.975 ($\pm$0.15)& 0.98 ($\pm$0.14) \\ \hline
crossfire &	0.08 	($\pm$ 0.27) & 0.05 (0.21) & 0.05 ($\pm$0.21) & 0.06 ($\pm$0.23)\\ \hline 
digdug &	0.00 ($\pm$ 0.00) & 0.00 ($\pm$ 0.00) & 0.00 ($\pm$ 0.00) & 0.00 ($\pm$ 0.00) \\ \hline
escape &	0.38 	($\pm$ 0.48) & 0.2 ($\pm$0.40) & 0.175 ($\pm$0.37)& 0.315 ($\pm$0.46)\\ \hline
hungrybirds &	0.30 ($\pm$ 0.45) & 0.27 ($\pm$0.44) & 0.33 ($\pm$0.47) & 0.285 ($\pm$0.45)\\ \hline
infection &	0.96 ($\pm$ 0.18) & 0.985 ($\pm$0.12) &0.995 ($\pm$0.07) & 0.955 ($\pm$0.20)\\ \hline
intersection &	1.00 ($\pm$ 0.00) &1.00 ($\pm$ 0.00)& 1.00 ($\pm$ 0.00) & 1.00 ($\pm$ 0.00) \\ \hline
lemmings &	0.00 ($\pm$ 0.00) & 0.00 ($\pm$ 0.00)& 0.00 ($\pm$ 0.00)& 0.00 ($\pm$ 0.00) \\ \hline
missilecommand	& 0.72 ($\pm$ 0.44) & 0.64 ($\pm$0.48) & 0.74 ($\pm$0.43) & 0.545 ($\pm$0.49)\\ \hline
modality	& 0.26 ($\pm$ 0.44) & 0.255 ($\pm$0.43) & 0.265 ($\pm$0.44) &0.26 ($\pm$0.43) \\ \hline
plaqueattack &	0.97 ($\pm$ 0.17) & 0.92 ($\pm$0.27)& 0.965 ($\pm$0.18) & 0.945 ($\pm$0.22)\\ \hline
roguelike &	0.00 ($\pm$	0.00) & 0.00 ($\pm$	0.00) & 0.00 ($\pm$	0.00) & 0.00 ($\pm$	0.00) \\ \hline
seaquest	&0.88 ($\pm$ 0.32) &	0.75 ($\pm$0.43) & 0.87 ($\pm$0.33) & 0.88 ($\pm$0.32)\\  \hline
survivezombies	& 0.50 ($\pm$ 0.49)& 0.415 ($\pm$0.49)& 0.455 ($\pm$0.49) & 0.475 ($\pm$0.49)\\ \hline
waitforbreakfast & 0.75 ($\pm$ 0.43) &0.67 ($\pm$0.47) & 0.69 ($\pm$0.46) & 0.64 ($\pm$0.48)\\ \hline

\end{tabular}
}
\label{tab:detailsWins}
\end{table}

\end{document}